\documentclass{bmvc2k}

\usepackage{booktabs}
\usepackage{multirow}
\usepackage{amsmath}
\usepackage{amssymb}
\usepackage{xcolor}
\usepackage{wrapfig}

\usepackage{xspace}
\usepackage{rotating}
\usepackage{pdflscape}
\newcommand*\rot{\rotatebox{75}}
\newcommand{\bb}[1]{\textbf{#1}}
\usepackage{array}
\newcolumntype{C}{>{\centering\arraybackslash}p{0.5em}}
\newcolumntype{D}{>{\centering\arraybackslash}p{0.8em}}

\def\x2{\chi^2}

\def\Re{\mathbb R}

\def\1{\mathds{1}}

\DeclareMathOperator*{\argmin}{\arg\!\min}

\usepackage{times}
\usepackage{epsfig}
\usepackage{graphicx}
\usepackage{moresize}
\usepackage{placeins}

\usepackage{algorithmic}
\usepackage{algorithm}

\title{Deep Fishing}

\addauthor{Albert Gordo}{albert.gordo@xrce.xerox.com}{1}
\addauthor{Adrien Gaidon}{adrien.gaidon@xrce.xerox.com}{1}

\addauthor{Florent Perronnin}{perronnin@fb.com}{2}

\addinstitution{
Computer Vision Group\\
Xerox Research Centre Europe 
}
\addinstitution{
Facebook AI Research$^*$\\
}

\runninghead{Gordo \etal}{Deep Fishing: Gradient Features from Deep Nets}

\def\eg{\emph{e.g}\bmvaOneDot}
\def\ie{\emph{i.e}\bmvaOneDot}

\def\cf{\emph{cf}\bmvaOneDot}

\def\etal{\emph{et al}\bmvaOneDot}
\def\wrt{\emph{w.r.t}\bmvaOneDot}

\makeatletter
\def\blfootnote{\xdef\@thefnmark{}\@footnotetext}
\makeatother
\begin{document}
\blfootnote{$^*$ Work done while FP was at the Computer Vision Group of the Xerox Research Centre Europe.}

\title{\Large Deep Fishing: Gradient Features from Deep Nets}
\maketitle

\begin{abstract}
Convolutional Networks (ConvNets) have recently improved image recognition performance thanks to
end-to-end learning of deep feed-forward models from raw pixels.
Deep learning is a marked departure from the previous state of the art, the Fisher Vector (FV),
which relied on gradient-based encoding of local hand-crafted features.
In this paper, we discuss a novel connection between these two approaches. First, we show that one
can derive gradient representations from ConvNets in a similar fashion to the FV.
Second, we show that this gradient representation actually corresponds to a structured matrix
that allows for efficient similarity computation.
We experimentally study the benefits of transferring this representation over the outputs of
ConvNet layers, and find consistent improvements on the Pascal VOC 2007 and 2012 datasets.
\end{abstract}
\section{Introduction}
\label{sec:intro}

Image classification involves describing images with pre-determined labels.
One of the first breakthroughs towards solving this problem was the
bag-of-visual-words (BOV)~\cite{sivic03bov,csurka04bov}.
While the BOV simply involves counting the number of occurrences of quantized
local features, approaches that encode higher order statistics such as the the Fisher Vector (FV)~\cite{perronnin2007fv, perronnin2010improving} led to state-of-the-art image classification results~\cite{chatfield2011devil,
sanchez2013fv}. Especially, such higher-order encodings were used by the leading
teams in the 2010 and 2011 editions of the ImageNet Large Scale Visual
Recognition Challenge (ILSVRC)~\cite{deng2009imagenet,
russakovsky2014imagenet}.
FV-based approaches were however outperformed in 2012 by the work of Krizhevsky
\etal~\cite{krizhevsky2012cnn} based on Convolutional Networks
(ConvNets)~\cite{lecun1989handwritten} trained in a supervised fashion on large
amounts of labeled data.
These models are feed-forward architectures involving multiple computational
layers that alternate linear operations, \eg convolutions, and non-linear
operations, \eg rectified linear units (ReLU). 
The end-to-end training of the large number of parameters inside ConvNets from
pixels to the specific end-task is a key to their success.
Since then, ConvNets, including improved architectures~\cite{Zeiler2014,
sermanet2014overfeat, simonyan2014verydeep}, have consistently outperformed all
other alternatives in subsequent editions of ILSVRC.
Also, ConvNets have remarkable \emph{transferability properties} when
used as ``universal'' feature extractors~\cite{yosinski2014}: if one feeds an
image to a ConvNet, the output of intermediate layers might be used as a
representation of this image and typically fed to linear classifiers. To the
best of our knowledge, this heuristic is not based on a strong theoretical
ground, but has been experimentally shown to work well in
practice~\cite{donahue2014decaf, oquab2014transferring, Zeiler2014,
chatfield2014return, Razavian2014}.

\begin{figure*}
\includegraphics[width=\textwidth]{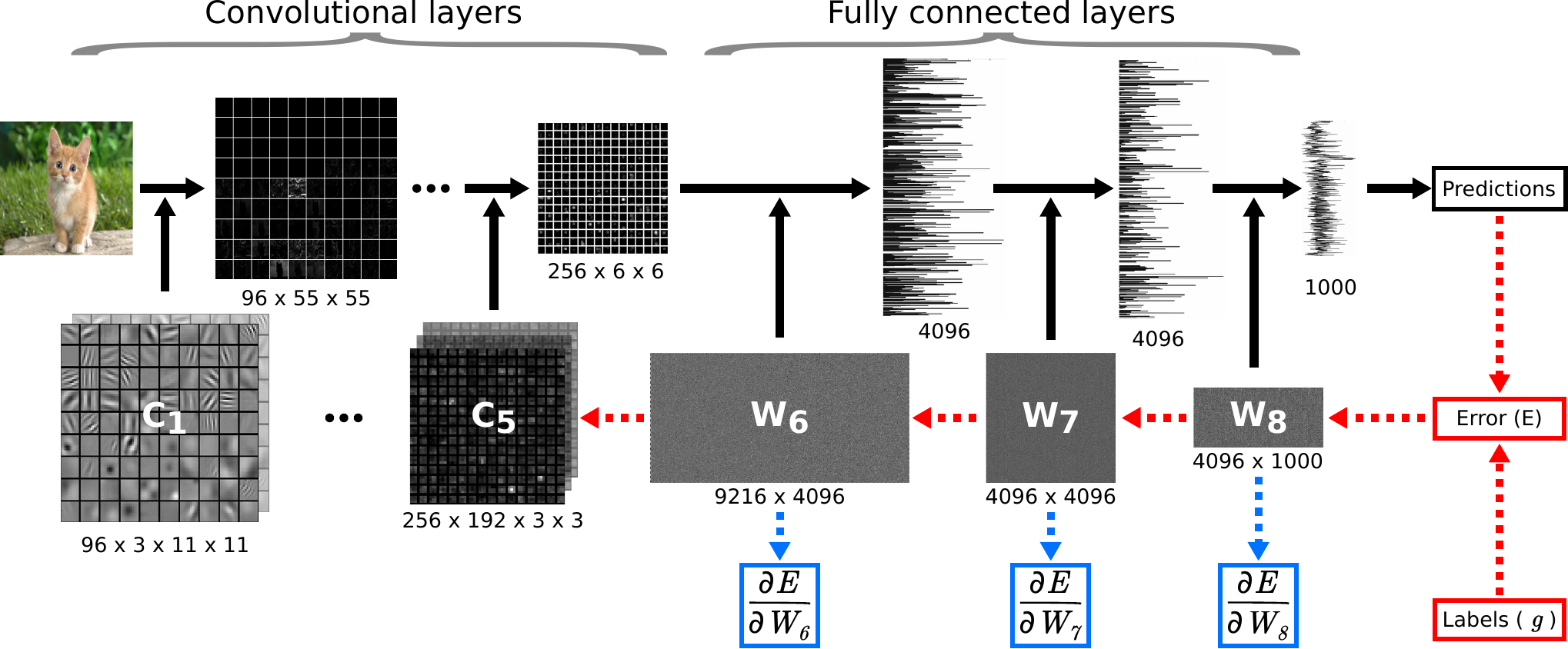}
\vspace*{1mm}
\caption{\small AlexNet architecture~\cite{krizhevsky2012cnn}. $C_k$ are the
parameters (4D tensors) of the convolutional layers.  $W_k$ are the parameters
(matrices) of the fully connected layers.  Black (resp. red) arrows represent
the information flow during the forward (resp. backward) pass.  Inspired by the
Fisher Kernel~\cite{jaakkola1998fk}, we study the use of gradient-related
information (the blue matrices) as transferable representations.}
\label{fig:overview}
\vspace*{-3mm}
\end{figure*}

Although ConvNets and FV approaches differ significantly, several works tried
to combine their benefits~\cite{simonyan2013deep, sydorov2014deep,
gong2014multiscale, perronnin2015}. Our work also attempts to \emph{get the best of both FV
and ConvNet worlds}.
Our {\bf primary contribution} is a novel approach to extract a
\emph{transferable} representation of an image given a pre-trained ConvNet. We
draw inspiration from the FV, which is based on the theoretically well-founded
Fisher Kernel (FK) proposed by Jaakkola and Haussler~\cite{jaakkola1998fk}. The
FK involves deriving a kernel from an underlying generative model of the data
by taking the gradient of the log-likelihood with respect to the model
parameters.  In a similar manner, given an unlabeled image, we propose to
compute the \emph{gradient of a cross-entropy criterion measured between the
predicted class probabilities and an equal probability output}.
This gradient with respect to the parameters of the fully connected layers
yields very high-dimensional representations (\cf Figure~\ref{fig:overview}).
Our {\bf second contribution} consists in leveraging the special structure of
this gradient representation to design an efficient kernel. We show that our
representation actually corresponds to a rank-1 matrix, for which the trace
kernel can be efficiently computed. Furthermore, this kernel decomposes in our
case into the product of two simpler kernels: the standard one on forward-pass
features, and a second one on quantities efficiently computed by
back-propagation.

The remainder of this article is organized as follows. In
section~\ref{sec:related}, we review related works. In section~\ref{sec:back},
we provide more background on the FK and ConvNets.  In section
~\ref{sec:fisherizing}, we introduce our novel hybrid ConvNet-gradient
representation as well as our associated efficient kernel. Finally, we provide
experimental results on the PASCAL VOC 2007 and 2012 benchmarks in
section~\ref{sec:exp}, showing that our representation consistently transfers
better than the standard forward pass features.
\section{Related Work}
\label{sec:related}

{\bf Hybrid techniques.} 
%
Several works have proposed to combine the benefits of deep learning with
"shallow" bag-of-patches representations based on higher-order statistics such
as the FV~\cite{perronnin2007fv, perronnin2010improving} or the
VLAD~\cite{jegou2010vlad}.
Simonyan~\etal~\cite{simonyan2013deep} propose to stack multiple FV layers,
each defined as a set of five operations: i) FV encoding, ii) supervised
dimensionality reduction, iii) spatial stacking, iv) $\ell_2$ normalization and
v) PCA dimensionality reduction. They show that, when combined with the
original FV, such networks lead to significant performance improvements on
ImageNet.
Peng~\etal~\cite{peng2014action} proposed a similar idea, but for action recognition.
Alternatively, Sydorov~\etal~\cite{sydorov2014deep} improve on the FV framework
by jointly learning the SVM classifier and the GMM visual vocabulary.
Conceptually, this is similar to back-propagation as used to learn neural
network parameters: the gradients corresponding to the SVM layer are
back-propagated to compute the gradients with respect to the GMM parameters.
Peng~\etal~\cite{peng2014vlad} proposed a similar idea for the VLAD~\cite{jegou2010vlad} descriptor.
Finally, Gong~\etal~\cite{gong2014multiscale} address the lack of geometric
invariance in ConvNets with a hybrid approach. They extract mid-level ConvNet
features from large patches, embed them using the VLAD encoding, and aggregate
them at multiple scales. This leads to competitive results on a number of
classification tasks.
While our goal -- getting the best of the FV and deep frameworks -- is shared
with these previous works, we differ significantly, as we are the first to
propose to derive gradient features from deep nets.

\noindent{\bf Deriving representations from pre-trained classifiers.}
%
Classemes~\cite{wang2009flickr, torresani2010classemes} is a common image
representation from a set of classifiers obtained by simply stacking classifier
scores.
Dimensionality reduction is generally applied on classeme
features~\cite{douze2011combining}, but learning separately the classification
and dimensionality reduction is suboptimal~\cite{gordo2012leveraging}.
Several works~\cite{bergamo2011picodes, weston2010wsabie, gordo2012leveraging}
learn an optimal embedding of images in a low-dimensional space via classifiers
with an intermediate hidden layer.  The first layer can be understood as a
supervised dimensionality reduction step, while the second one can be
interpreted as a set of classifiers in the intermediate space. A new image is
represented as the output of this intermediate layer, discarding the
classifiers.
A natural extension is to learn deeper architectures, \ie architectures with
more than one hidden layer, and to use the output of these intermediate layers
as features for the new tasks.  Krizhevsky \etal~\cite{krizhevsky2012cnn}
proposed to learn end-to-end a deep classifier based on the ConvNet
architecture of LeCun \etal~\cite{lecun1989handwritten}. They showed
qualitatively that the output of the penultimate layer could be used for image
retrieval.  This finding was quantitatively validated for a number of tasks,
including image classification~\cite{donahue2014decaf, oquab2014transferring,
Zeiler2014, chatfield2014return, Razavian2014}, image
retrieval~\cite{Razavian2014, Babenko2014}, object
detection~\cite{girshick2014rich}, and action
recognition~\cite{oquab2014transferring}.
The choice of the layer(s) whose output should be used for representation
purposes depends on the problem at hand. As observed by Yosinski
\etal~\cite{yosinski2014}, this choice should be driven by the distance between
the base task (the one used to learn the classifier) and target task.  
In this paper, we show that this heuristic of using the output of an
intermediate-level fully connected layer as image representation can be related
to the application of the Fisher Kernel idea to ConvNets.
\section{Background on the Fisher Kernel and ConvNets}
\label{sec:back}

\subsection{Fisher Kernel}

The Fisher Kernel (FK) is a generic principle introduced to combine the benefits of generative and
discriminative models to pattern recognition.  Let $X$ be a sample, and let $u_{\theta}$ be a
probability density function that models the generative process of $X$, where $\theta$ denotes the
vector of parameters of $u_{\theta}$.  In statistics, the {\em score function} is given by the
gradient of the log-likelihood of the data on the model:
\begin{equation}
\varphi^{FK}_{\theta}(X) = \nabla_{\theta} \log u_{\theta}(X) .
\end{equation}
This gradient describes the contribution of the individual parameters to the generative process.
Jaakkola and Haussler \cite{jaakkola1998fk} proposed to measure the similarity between two samples
$X$ and $Y$ using the {\em Fisher Kernel} (FK) which is defined as:
\begin{equation} \label{eqn:fk}
K_{FK}(X,Y) = {\varphi^{FK}_{\theta}(X)}' F_{\theta}^{-1} \varphi^{FK}_{\theta}(Y)
\end{equation}
where $F_{\theta}$ is the Fisher Information Matrix, usually approximated by the identity matrix~\cite{jaakkola1998fk}.
One of the benefits of the FK framework is that it comes with guarantees.  The FK is indeed
asymptotically at least as good as the MAP decision rule, when assuming that the classification
label is included in the generative model as a latent variable (theorem 1
in~\cite{jaakkola1998fk}).
Some extensions make the dependence of the kernel on the classification labels
explicit.  This includes the likelihood kernel~\cite{fine2001hybrid}, which
involves one generative model per class, and which consists in computing one FK
for each generative model (and consequently for each class). This also includes
the likelihood ratio kernel~\cite{smith2001speech}, which is tailored to the
two-class problem, and which involves computing the gradient of the
log-likelihood of the ratio between the two class likelihoods. Given two
classes denoted $c_1$ and $c_2$ with class-conditional probability density
functions $p(.|c_1)$ and $p(.|c_2)$ and with collective parameters $\theta$,
this yields:
\begin{equation}
\label{eqn:lr}
\varphi^{LR}_\theta (X) = \nabla_{\theta} \log \frac{p(X|c_1)}{p(X|c_2)}.
\end{equation}
The likelihood ratio kernel is supported by strong experimental
evidence~\cite{smith2001speech} and theory~\cite{smith2002svms}.
In section \ref{sec:fisherizing}, we extend it to derive a gradient
representation from a ConvNet model.

\subsection{Convolutional Networks}
\label{s:convnets}

Convolutional Networks (ConvNets)~\cite{lecun1989handwritten} are the de facto state-of-the-art
models for image recognition since the work of Krizhevsky \etal~\cite{krizhevsky2012cnn}.
This class of deep learning models relies on a feed forward architecture typically composed of a
stack of convolutional layers followed by a stack of fully connected layers (see
Figure~\ref{fig:overview} for the standard AlexNet~\cite{krizhevsky2012cnn} architecture).
A convolutional layer is parametrized by a 4D tensor representing a stack of 3D filters. During the
forward pass, these filters are run in a sliding window fashion across the output of the previous
layer (or the image itself for the first layer) in order to produce a 3D tensor: the stack of
per-filter activation maps. These activation maps then pass through a non-linearity (typically a
Rectified Linear Unit, or ReLU~\cite{krizhevsky2012cnn}) and an optional pooling stage before being
fed to the next layer.
Both the standard AlexNet~\cite{krizhevsky2012cnn} and recent improved
architectures like VGGNet~\cite{simonyan2014verydeep} use a stack of fully
connected layers to transform the output activation map of the convolutional
layers into class-membership probabilities.
A fully connected layer consists in a simple matrix vector multiplication
followed by a non-linearity, typically ReLU for intermediate layers and a
SoftMax for the last one.

Let $x_k$ be the output of layer $k$, which is also the input of layer $k+1$
(for AlexNet, $x_5$ is the flattened activation map of the fifth convolutional
layer).
Layer $k$ is parametrized by the 4D tensor $C_k$ if it is a convolutional
layer, and by the matrix $W_k$ for a fully connected layer. A fully connected
layer performs the operation $x_k = \sigma(W_k^T x_{k-1})$, where $\sigma$ is
the non-linearity. We note $y_k = W_k^T x_{k-1}$ the output of layer $k$ before
the non-linearity, and $\theta = \{C_1, \cdots, C_M\} \cup \{W_{M+1}, \cdots,
W_L\}$ the parameters of all $L$ layers of the network.
Training such deep models consists in end-to-end learning of this vast number
of parameters via the minimization of an error (or loss) function on a large
training set of $N$ image and ground-truth label pairs $(I^i, g^i)$. The
typical loss function used for classification is the cross-entropy:
\begin{equation}
E(I^i, g^i; \theta) = - \sum_{j=1}^P g^i_j \log(x^i_{L, j})
\label{eq:error}
\end{equation}
where $P$ is the number of labels (categories), $g^i \in \{0,1\}^P$ is the
label vector of image $I^i$, and $x^i_{L, j}$ is the predicted probability of
class $j$ for image $I^i$ resulting from the forward pass.

\noindent
The optimal network parameters $\theta^*$ are the ones minimizing the loss over
the training set:
\begin{equation}
\theta^* = \argmin_\theta \sum_{i=1}^N E(I^i, g_i; \theta)
\label{eq:objective}
\end{equation}
This optimization problem is typically solved using Stochastic Gradient Descent
(SGD)~\cite{bottouSGD}, a stochastic approximation of batch gradient descent consisting in doing
approximate gradient steps equal on average to the true gradient $\nabla_\theta E$.
Each approximate gradient step is typically performed with a small batch of labeled examples
in order to efficiently leverage the caching and vectorization mechanisms of modern
hardware.

A particularity of deep networks is that the gradients with respect to all
parameters $\theta$ can be computed efficiently in a stage-wise fashion via a
sequential application of the chain rule
(``back-propagation''~\cite{lecun1989handwritten}).
In particular, when using ConvNets as feature extractors, the first phase consists in pre-training
the network (\ie obtaining $\theta^*$) via SGD with back-propagation on a large labeled dataset like
ImageNet~\cite{russakovsky2014imagenet}.
Then, ConvNet features can be used for different tasks using forward passes on the pre-trained
network.
In the following, we describe how we can also use back-propagation \emph{at test time} to transfer
richer Fisher Vector-like representations based on the gradient of the loss with respect to the
ConvNet parameters.

\section{Gradient Features from Deep Nets}
\label{sec:fisherizing}
We now motivate the use of gradient features from deep nets by relating the
likelihood ratio kernel in equation~\eqref{eqn:lr} to the ConvNet objective
function in equation~\eqref{eq:error}.  We then explicit the gradient equations
and relate our gradient features to the standard heuristic features derived
from the outputs of intermediate layers.  Finally, we explain how to
efficiently compute  the similarity between these high-dimensional
representations.

\subsection{Relating the likelihood ratio kernel and deep nets}
\label{sec:obj}
The FK~\cite{jaakkola1998fk} and its
extensions~\cite{fine2001hybrid,smith2001speech} were proposed as generic
frameworks to derive representations and kernels from generative models.  As
the standard ConvNet classification architecture does not define a generative
model, such frameworks cannot be applied as-is. However, we can draw
inspiration from the likelihood ratio kernel for that purpose.
We start from equation \eqref{eqn:lr} and note that it can be rewritten as the
gradient of the log-likelihood of the ratio between posterior probabilities
(assuming equal class priors), \ie:
\begin{equation}
\label{eqn:pr}
\varphi^{LR}_\theta (X) = \nabla_{\theta} \log \frac{p(X|c_1)}{p(X|c_2)} = \nabla_{\theta} \log \frac{p(c_1|X)}{p(c_2|X)}.
\end{equation}
In the two-class problem of~\cite{smith2001speech}, we have $p(c_2|X) = 1 - p(c_1|X)$ and equation
$\eqref{eqn:pr}$ gives:
\begin{equation}
\label{eqn:prj}
\varphi^{LR}_\theta (X)
= \frac{\varphi^{pos}_{\theta;1} (X)}{1 - p(c_1|X)}
\end{equation}
where $\varphi^{pos}_{\theta;j}=\nabla_\theta \log p(c_j|X)$ 
is the gradient of the log-posterior for class $c_j$.
We underline that the previous formula is general in the sense that
it can be applied beyond generative models.
To extend this representation beyond the two-class case, one may compute an
embedding $\varphi^{pos}_{\theta;j}$ for each class $j$ using the gradient of
the corresponding log-posterior probability.

We can now observe the relation between the ConvNet objective $E$ in equation \eqref{eq:error} 
for an image $I$ and label vector $g$ with these gradient of log-posterior embeddings:

\begin{equation}
\label{eq:fisherizing}
\nabla_\theta E(I, g; \theta)
= - \sum_{j=1}^P g_j \nabla_{\theta} \log p(c_j | I) 
= - \sum_{j=1}^P g_j \varphi^{pos}_{\theta;j}(I)
\end{equation}
Consequently, the gradient of the ConvNet objective can be interpreted as a
sum of gradient embeddings $\varphi^{pos}_{\theta;j}(I)$, weighted by the labels $g_j$.

To use this gradient as an image representation, as is the case of the FK,
there are two main challenges to be addressed.
First, we do not have access to the value of the label $g$, which we
need to compute the representation of a test image $I$ according to
equation~\eqref{eq:fisherizing}.
The simplest solution consists in using a constant uniform label vector $g =
\bar{g} =[1/P, \ldots, 1/P]$. Although $\bar{g}$ is non-informative, we
experimentally validate the interest of this simple strategy.
The second issue concerning the use of $\nabla_\theta E(I, \bar{g}; \theta)$ as
an image representation lies in the associated computational cost.
Although scalable in the number of classes, this representation is very
high-dimensional. The number of parameters $\theta$ in current deep
architectures is indeed too large to be able to use the full gradient
$\nabla_\theta E$ in practice.
Therefore, we propose to use \emph{only the partial derivatives with respect to
the parameters of some fixed layers}, in the same spirit as what is currently
done with layer-activation features.
These partial derivatives can be computed and compared efficiently using the
chain rule and a rank 1 decomposition, as shown in the following sections.
Note also that this approach can be further combined with other existing
techniques, including ones specialized for deep nets (\eg model
compression~\cite{bucilua2006}) or for FV (\eg~product
quantization~\cite{jegou2011pq}).

\subsection{Gradient derivation}
\label{sec:grad}

One remarkable property of ConvNets and other feed-forward architectures is that they are
differentiable through all their layers.  In the case of ConvNets, it is easy to show that the
gradients of the loss with respect to the weights of the fully-connected layers are:
\begin{equation}
\frac{\partial E}{\partial W_k} = x_{k-1} \left[\frac{\partial E}{\partial y_k}\right]^T.
\label{eq:weight_derivative}
\end{equation}

To compute the partial derivatives of the loss with respect to the output parameters needed in
Equation \eqref{eq:weight_derivative}, one can apply the chain rule.
In the case of fully-connected layers and ReLU non-linearities, this leads to the following
recursive definition and base case,
\begin{align}
\frac{\partial E}{\partial y_k}  = 
    \left[ W_{k+1} \frac{\partial E}{\partial y_{k+1}}\right] \circ  \mathbb{I}_{[y_k > 0]}, &\hspace{1cm}& \frac{\partial E}{\partial y_L} =  \bar{g} - \sigma(y_L),
\label{eq:y_derivative}
\end{align}
where $\mathbb{I}_{[y > 0]}$ is an indicator vector, set to one at the
positions where $y > 0$ and to zero otherwise, $\circ$ is the Hadamard or
element-wise product, $\bar{g}$ is a supplied vector of labels with which
to compute the loss, and $\sigma$ is the SoftMax function. From the previous
section, we use $\bar{g}=[1/P, ..., 1/P]$, \ie we assume that all classes have
equal probabilities.
It is worth noticing how $\frac{\partial E}{\partial y_L}$ is simply a shifted version of the output
probabilities, while the derivatives \wrt $y_{k}$ with $k<L$ are linear transformations of these
shifted probabilities, as the Hadamard product can be rewritten as a matrix multiplication.

\subsection{Computing similarities between gradients}
\label{sec:sim}

Using the gradients in equation~\eqref{eq:weight_derivative} as features is problematic in practice
due to their high-dimensional nature with current deep architectures. In the case of
AlexNet~\cite{krizhevsky2012cnn}, $\frac{\partial E}{\partial W_8}$ is around $4$ million floating
point values, while $\frac{\partial E}{\partial W_7}$ and $\frac{\partial E}{\partial W_6}$ are each
around $16$ and $36$ million floats.
Thus, explicitly computing the dot-product between the gradients is impractical. Instead, we
propose to take advantage of the unique structure of our gradients (rank-1
matrices, \cf~Eq.~\eqref{eq:weight_derivative}) by using the trace kernel, defined for two matrices $A$ and $B$ as:
\begin{equation}
K_{tr}(A,B)= Tr(A^TB)
\end{equation}
For rank-1 matrices, the trace can be decomposed as the product of two kernels. If we let $A=a
u^T$, $A \in \Re^{d\times D}$, and $B = b v^T$, $B \in \Re^{d\times D}$, with  $a$, $b \in \Re^d$
and $u$, $v \in \Re^D$, then:
\begin{equation}
    K_{tr}(A,B)  =  Tr(a u^T(b v^T)^T)  =  Tr(a u^T v b^T)\nonumber =  Tr(b^T a u^T v)     =  (b^Ta)\cdot (u^Tv).
\end{equation}
Therefore, for two images $A$ and $B$, we can compute the similarity between
gradients in a low-dimensional space without explicitly computing the gradients
\wrt the weights $W_k$:
\begin{equation}
K_{tr}  \left(
\frac{\partial E}{\partial W_k}(A, \bar{g}; \theta),
\frac{\partial E}{\partial W_k}(B, \bar{g}; \theta)
\right)
= 
(x_{k-1}^{A})^T x_{k-1}^{B}
\cdot
\left[\frac{\partial E}{\partial y_k}(A, \bar{g}; \theta)\right]^T 
\frac{\partial E}{\partial y_k}(B, \bar{g}; \theta)
\label{eq:tr}
\end{equation}
The left part of this equation indicates that the forward activations of the
two inputs should be similar. This is the standard measure of similarity which
is used between images when described by the outputs of the intermediate layers
of ConvNets.  However, this similarity is multiplicatively weighted by the
similarity between the back-propagated quantities. This indicates that, to
obtain a high value with the proposed kernel, both the target forward
activations \emph{and} the back-propagated quantities of the images need to be
similar.

\noindent \textbf{Normalization.}
The $\ell_2$-normalization of the activation features consistently leads to superior results~\cite{chatfield2014return}.
In our experiments we $\ell_2$-normalize our forward and backward features independently.
This is consistent with normalizing the gradient matrix using a Frobenius norm, since $||au^T||_{F} = ||a||_2 ||u||_2$.

\section{Experimental results}
\label{sec:exp}
\subsection{Datasets and evaluation protocols}

We evaluate our approach to transfer features from pretrained models on two
standard classification benchmarks, Pascal VOC 2007 and Pascal VOC 2012
\cite{pascal-voc-2007}.  These datasets contain $9,963$ and $22,531$ annotated
images, respectively. Each image is annotated with one or more labels
corresponding to $20$ object categories. The datasets include partitions for
training, validating, and testing, and the accuracy is measured in terms of per
class mean average precision (mAP).
The test annotations of VOC 2012 are not public, but an evaluation server with a
limited number of submissions per week is available. Therefore, we use the
validation set for the first part of our analysis on the VOC 2012 dataset, and
evaluate on the test set only for the final experiments.
We conduct all VOC 2007 experiments on the full dataset.

\subsection{Implementation details}

We tested our approach on two different deep ConvNets:
AlexNet~\cite{krizhevsky2012cnn} and VGG16~\cite{simonyan2014verydeep}.  VGG16
is a much deeper architecture than AlexNet, with many more convolutional
layers, leading to superior performance, but also to a much slower training and
feature extraction.  We used the pre-trained networks that are publicly
available\footnote{\url{https://github.com/BVLC/caffe/wiki/Model-Zoo}}. Both
networks were pre-trained on the ILSVRC2012 subset of ImageNet, which is
disjoint from the Pascal VOC datasets, and therefore suitable for our
evaluation of feature transfer.

To extract descriptors from the Pascal images, we first resize the images so
that the shortest size has $227$ pixels ($224$ on the VGG16 case), and then
take the central square crop, without distorting the aspect ratio.  We found
this cropping technique to work well in practice. For simplicity, we do no data
augmentation.
The feature extraction is performed on a customized version of the \emph{caffe}
library\footnote{\url{http://caffe.berkeleyvision.org}}, modified to expose the
back-propagation features.  This allows us to extract forward and backward
features of the training and testing images.  At testing time we use a tempered
version of SoftMax, $\sigma(y,\tau) = exp(y/\tau)/\sum_i exp(y_i/\tau)$, with
$\tau=2$, to produce softer probability distributions for backpropagation.  As
discussed in section~\ref{sec:obj}, we use non-informative uniform labels for
the backward pass to extract the gradient features.  All forward and backward
features are then $\ell_2$-normalized. 

To perform classification, we use the SVM implementation of
\emph{scikit-learn}~\cite{scikit-learn}\footnote{\url{http://scikit-learn.org/}}.
The cost parameter $C$ of the solver was set to the default value of $1$, which
worked well in practice.

\begin{table}[!tb]
    \caption{\small Left: Results on Pascal VOC 2007 and VOC 2012 with AlexNet
(A) and VGG16 (V). Results on VOC 2012 are on the validation set. Right:
Comparison with other ConvNet results (mAP in \%).}
\vspace*{2mm}
    \begin{minipage}{.45\linewidth}
\vspace{0.1cm}
\centering
\footnotesize
\begin{tabular}{l|DD|DD}
&  \multicolumn{2}{c}{VOC2007} & \multicolumn{2}{c}{VOC2012}  \\
Features & (A) & (V) & (A) & (V) \\
\midrule
$x_5$ (Pool5) & 71.0 & 86.7 & 66.1 & 81.4\\
$x_6$ (FC6) & 77.1 & 89.3 & 72.6 & 84.4\\
$x_7$ (FC7) & 79.4 & 89.4 & 74.9 & 84.6\\
$y_8$ (FC8) & 79.1 & 88.3 & 74.3 & 84.1\\
$x_8$ (Prob) & 76.2 & 86.0 & 71.9 & 81.3\\
\midrule
$x_5$; $x_6$ & 76.4 & 89.2 & 71.6 & 84.0\\
$\frac{\partial E}{\partial W_6} =  x_5 \left[\frac{\partial E}{\partial y_6}\right]^T$ & 80.2 & 89.3 & 75.1 & 84.6\\ 
\midrule
$x_6$; $x_7$ & 79.1 & 89.5 & 74.3 & 84.6\\
$\frac{\partial E}{\partial W_7} = x_6 \left[\frac{\partial E}{\partial y_7}\right]^T$ &  \bb{80.9} & \bb{90.0} & \bb{76.3} & \bb{85.2}\\
\midrule
$x_7$; $y_8$ & 79.7 & 89.2 & 75.3 & 84.6\\
$\frac{\partial E}{\partial W_8} = x_7 \left[\frac{\partial E}{\partial y_8}\right]^T$ & 79.7 & 88.2 & 75.0 & 83.4\\
\bottomrule
\end{tabular}
\label{tab:summaryVocDual}
    \end{minipage}%
    \begin{minipage}{.5\linewidth}
\centering
\footnotesize
\begin{tabular}{ll|c|c}
\multicolumn{2}{c|}{} & VOC'07 & VOC'12 \\
\hline 
\multicolumn{2}{l|}{Proposed  - AlexNet~\cite{krizhevsky2012cnn}}  & 80.9 &  76.5 \\
\multicolumn{2}{l|}{Proposed -  VGG16~\cite{simonyan2014verydeep}}  & {\bf 90.0} & 85.3 \\
\hline 
DeCAF & \cite{donahue2014decaf} from \cite{chatfield2014return} & 73.4 & -\\
Razavian \etal  & \cite{Razavian2014} & 77.2 & - \\
Oquab \etal  & \cite{oquab2014transferring} & 77.7 & 78.7 \\
Zeiler \etal  & \cite{Zeiler2014} & - & 79.0 \\
Chatfield \etal & \cite{chatfield2014return} & 82.4 &  83.2 \\
He \etal & \cite{He2014} & 80.1 & - \\
Wei \etal  & \cite{Wei2014} & 81.5 & 81.7 \\
Simonyan \etal &  \cite{simonyan2014verydeep} & 89.7 & {\bf 89.3} \\
\hline 
\end{tabular}
\label{tab:pascal_sota}

    \end{minipage} 
\end{table}

\subsection{Results and discussion}

Table~\ref{tab:summaryVocDual} summarizes the results, and compares our
approach with the state of the art and different baselines.
We extract and compare several features for each dataset and network
architecture: (i) individual forward activation features, from Pool5 up to the
probability layer; (ii) concatenation of forward activation features, \eg
Pool5+FC6, FC6+FC7, FC7+FC8; (iii) our proposed gradient features:
$\frac{\partial E}{\partial W_6}$, $\frac{\partial E}{\partial W_7}$, and
$\frac{\partial E}{\partial W_8}$.
The similarity between $\ell_2$-normalized forward activation features is
measured with the dot-product, while the similarity between gradient
representations is measured using the trace kernel.  We highlight the following
points.

\textbf{Forward activations.}
In all cases, FC7 is the best performing individual layer on both VOC2007 and
VOC2012, independently of the network. This is consistent with previous
findings.  Also consistent is the fact that the probability layer performs
badly in this case.  More surprisingly, concatenating forward layers does not
seem to bring any noticeable accuracy improvements in any setup.

\textbf{Gradient representations.}
We compare the gradient representations with the concatenation of forward
activations, since they are very related and share part of the features.
On the deeper layers ($6$ and $7$) the gradient representations
outperform the individual features as well as the concatenation both for
AlexNet and VGG16 on both datasets.  For AlexNet, the improvements are quite
significant: $+3.8\%$ and $+3.5\%$ absolute improvement for the gradients with
respect to $W_6$ on VOC2007 and VOC2012, and $+1.8\%$ and $+2\%$ for $W_7$.
The improvements for VGG16 are more modest but still noticeable: $+0.1\%$ and
$+0.6\%$ for the gradients with respect to $W_6$ and $+0.5\%$ and $+0.6\%$ for
the gradients with respect to $W_7$.  Larger relative improvements on less
discriminative networks such as AlexNet seem to suggest that the more complex
gradient representation can, to some extent, compensate for the lack of
discriminative power of the network, but that one obtains diminishing returns
as the power of the network increases.
Once one reaches the top of the network ($FC8$), the gradient representations
perform worse and these improvements diminish or disappear completely. This is
expected, as the derivative with respect to $W_8$ depends heavily on the output
of the probability layer, which is known to saturate.  However, for the
derivatives with respect to $W_6$ and $W_7$, more information is involved,
leading to superior results.

\textbf{Comparison with other works.} 
Our best results are compared with the state-of-the-art on PASCAL VOC2007 and
VOC2012 in Table~\ref{tab:pascal_sota}.  We can see that we obtain competitive
performance on both datasets.  We note however that our results with VGG16 are
somewhat inferior to those reported in~\cite{simonyan2014verydeep} with a
similar model.  We believe this might be explained by the more costly feature
extraction strategy employed by Simonyan and Zisserman which involves
aggregating image descriptors at multiple scales.

\begin{table}[t!]
    \tiny
    \centering
    \caption{Results on Pascal VOC2007 with AlexNet and VGG16. Comparison between the standard forward activation features and the proposed gradient features.}
\vspace{0.5cm}
\begin{tabular}{ccCCCCCCCCCCCCCCCCCCCC}
Features & \rot{\textbf{mean}} & \rot{aeroplane} & \rot{bicycle} & \rot{bird} & \rot{boat} & \rot{bottle} & \rot{bus} & \rot{car} & \rot{cat} & \rot{chair} & \rot{cow} & \rot{diningtable} & \rot{dog} & \rot{horse} & \rot{motorbike} & \rot{person} & \rot{pottedplant} & \rot{sheep} & \rot{sofa} & \rot{train} & \rot{tvmonitor}\\
\midrule
& \multicolumn{21}{c}{AlexNet}\\
\midrule
$x_7$ (FC7) & 79.4  & 95.4 & 88.6 & 92.6 & 87.3 & 42.1 & 80.1 & 90.5 & 89.6 & 59.9 & 68.2 & 74.1 & 85.3 & 89.8 & 85.6 & 95.3 & 58.1 & \bb{78.9} & 57.9 & 94.7 & \bb{74.4}\\
$\frac{\partial E}{\partial W_7}$ & \bb{80.9} & \bb{96.6} & \bb{89.2} & \bb{93.8} & \bb{89.5} & \bb{44.9} & \bb{81.0} & \bb{91.9} & \bb{89.9} & \bb{61.2} & \bb{70.4} & \bb{78.5} & \bb{86.2} & \bb{91.4} & \bb{87.4} & \bb{95.7} & \bb{60.5} & 78.8 & \bb{62.5} & \bb{95.2} & 73.5\\ 
\midrule
& \multicolumn{21}{c}{VGG16}\\
\midrule
$x_7$ (FC7) & 89.3  & 99.2 & 95.9 & \bb{99.1} & 96.9 & \bb{63.8} & 92.8 & 95.1 & 98.1 & 70.4 & 87.8 & 84.3 & 97.0 & 97.2 & 93.5 & 97.3 & 68.6 & 92.2 & 73.3 & 98.7 & \bb{85.5}\\
$\frac{\partial E}{\partial W_7}$  & \bb{90.0} & \bb{99.6} & \bb{97.2} & 98.8 & \bb{97.0} & 63.3 & \bb{93.8} & \bb{95.6} & \bb{98.4} & \bb{71.1} & \bb{89.4} & \bb{85.3} & \bb{97.7} & \bb{97.7} & \bb{95.6} & \bb{97.5} & \bb{70.3} & \bb{92.7} & \bb{76.2} & \bb{98.8} & 84.2\\ 
\bottomrule
\end{tabular}
\label{tab:voc07}
\end{table}

\begin{table}[t!]
    \tiny
    \centering
	\caption{Results on Pascal VOC2012 with AlexNet and VGG16.  Comparison between the standard forward activation features and the proposed gradient features.}
\vspace{0.5cm}
\begin{tabular}{ccCCCCCCCCCCCCCCCCCCCC}
Features & \rot{\textbf{mean}} & \rot{aeroplane} & \rot{bicycle} & \rot{bird} & \rot{boat} & \rot{bottle} & \rot{bus} & \rot{car} & \rot{cat} & \rot{chair} & \rot{cow} & \rot{diningtable} & \rot{dog} & \rot{horse} & \rot{motorbike} & \rot{person} & \rot{pottedplant} & \rot{sheep} & \rot{sofa} & \rot{train} & \rot{tvmonitor}\\
\midrule
& \multicolumn{21}{c}{AlexNet (evaluated on the validation set)}\\
\midrule
$x_7$ (FC7) & 74.9  & 92.9 & 75.4 & 88.7 & 81.7 & 48.0 & 89.0 & 70.3 & 88.0 & 62.3 & 63.6 & 57.8 & 83.5 & 78.0 & 82.9 & 92.9 & 49.1 & 74.8 & 50.5 & 90.2 & 78.7\\ 
$\frac{\partial E}{\partial W_7}$& \bb{76.3} & \bb{94.3} & \bb{77.4} & \bb{89.5} & \bb{82.2} & \bb{50.8} & \bb{90.2} & \bb{72.4} & \bb{89.3} & \bb{64.8} & \bb{63.9} & \bb{60.3} & \bb{84.0} & \bb{79.6} & \bb{84.0} & \bb{93.2} & \bb{50.6} & \bb{76.7} & \bb{52.6} & \bb{91.8} & \bb{79.2}\\
\midrule
& \multicolumn{21}{c}{AlexNet (evaluated on the test set)}\\
\midrule
$x_7$ (FC7) & 75.0 & 93.8 & 75.0 & 86.4 & 82.2 & 48.2 & 82.5 & 73.8 & 87.6 & 63.8 & 63.5 & 69.3 & 85.7 & 80.3 & 84.1 & 92.3 & 47.4 & 72.2 & 51.8 & 88.1 & 72.5\\ 
$\frac{\partial E}{\partial W_7}$& \bb{76.5} & \bb{95.0} & \bb{76.6} & \bb{87.7} & \bb{82.9} & \bb{52.5} & \bb{83.4} & \bb{75.6} & \bb{88.6} & \bb{65.3} & \bb{65.4} & \bb{69.8} & \bb{86.5} & \bb{82.1} & \bb{85.1} & \bb{93.0} & \bb{48.2} & \bb{74.5} & \bb{57.0} & \bb{88.4} & \bb{73.0}\\
\midrule
& \multicolumn{21}{c}{VGG16 (evaluated on the validation set)}\\
\midrule
$x_7$ (FC7) & 84.6  & 98.2 & 88.3 & 94.6 & 90.5 & 66.0 & 93.6 & 80.5 & 96.4 & \bb{73.9} & 81.3 & 70.2 & 93.0 & 91.3 & 91.3 & 95.1 & \bb{56.3} & 87.7 & 64.2 & \bb{95.8} & 84.5\\
$\frac{\partial E}{\partial W_7}$&  \bb{85.2} & \bb{98.6} & \bb{89.4} & \bb{94.7} & \bb{91.5} & \bb{67.2} & \bb{94.0} & \bb{80.9} & \bb{96.8} & 73.7 & \bb{83.7} & \bb{71.9} & \bb{93.4} & \bb{91.6} & \bb{91.5} & \bb{95.4} & 56.0 & \bb{88.3} & \bb{65.2} & 95.5 & \bb{85.2}\\
\midrule
& \multicolumn{21}{c}{VGG16 (evaluated on the test set)}\\
\midrule
$x_7$ (FC7) & 85.0 & 97.8 & 85.2 & \bb{92.3} & 91.1 & 64.5 & \bb{89.7} & 82.2 & 95.4 & 74.1 & \bb{84.7} & \bb{81.1} & 94.1 & 93.5 & 91.9 & 95.0 & \bb{57.9} & 86.0 & 67.8 & \bb{95.2} & \bb{81.5}\\
$\frac{\partial E}{\partial W_7}$& \bb{85.3} & \bb{98.0} & \bb{86.0} & 91.7 & \bb{91.3} & \bb{65.7} & 89.6 & \bb{82.4} & \bb{95.5} & \bb{74.5} & 84.2 & 80.7 & \bb{94.3} & \bb{93.7} & \bb{92.2} & \bb{95.4} & 57.7 & \bb{87.2} & \bb{69.2} & \bb{95.2} & 81.4\\ 
\bottomrule
\end{tabular}
\label{tab:voc12}
\end{table}

\textbf{Per-class results.}
We report per-class results for Pascal VOC2007 on Table  \ref{tab:voc07} and
for VOC2012 on Table \ref{tab:voc12}.  We compare the best forward features
(individual FC7) with the best gradient representation ($\frac{\partial
E}{\partial W_7}$).  The results on VOC2007 are on the test set.  For VOC2012,
we report results both on validation and on test.
We observe how, on both networks and datasets, the results are consistently
better even when the improvements are not large.
For AlexNet, the gradient representation has the best performance on 18 out of
the 20 classes on VOC2007, and on all classes for VOC2012.  For VGG, the
gradient representation is the best one on 17 out of the 20 classes both on
VOC2007 and VOC2012 (validation).  The differences between validation and test
on VOC2012 are minimal. 
\section{Conclusions}

In this paper we show a link between ConvNets as feature extractors and Fisher
Vector encodings.  We have introduced a gradient-based representation for
features extracted with ConvNets inspired by the Fisher Kernel framework.  This
representation takes advantage of the high-quality features learned by ConvNets
on an end-to-end supervised manner, and of the discriminative power of
gradient-based representations.  We also presented an approach to compute
similarities between gradients in an efficient manner without computing
explicitly the high-dimensional gradient representations.  We show that this
similarity can be seen as a weighed version of the forward feature similarities
that takes into account not only the features themselves, but also information
back-propagated from the ConvNet objective.  We tested our approach on the
Pascal VOC2007 and VOC2012 benchmarks using two different popular deep
architectures, showing consistent improvements over using only the individual
forward activation features or their combination as it is standard practice.

{\small
\bibliography{egbib}
}

\end{document}